\documentclass{article}
\usepackage{ijcai19}

\usepackage[dvipdfmx]{color} 
\usepackage{times}
\usepackage{soul}
\usepackage{url}
\usepackage[hidelinks]{hyperref}
\usepackage[utf8]{inputenc}
\usepackage[small]{caption}
\usepackage{graphicx}
\usepackage{amsmath}
\usepackage{booktabs} 
\usepackage{algorithm}
\usepackage{algorithmic}
\renewcommand{\algorithmiccomment}[1]{\bgroup\hfill//~#1\egroup}
\usepackage[subrefformat=parens,format=hang]{subcaption} 
\usepackage{multirow}

\title{FAVAE: Sequence Disentanglement using Information Bottleneck Principle}

\author{
Masanori Yamada$^1$\footnote{Both authors equally contributed to this paper.}\and
Heecheol Kim$^2$\footnotemark[1]\and
Kosuke Miyoshi$^2$\And
Hiroshi Yamakawa$^{2,3}$\\
\affiliations
$^1$NTT Secure Platform Laboratories\\
$^2$Dwango Artificial Intelligence Laboratory\\
$^3$The Whole Brain Architecture Initiative, a specified non-profit organization\\
\emails
yamada0224@gmail.com,
h-kim@isi.imi.i.u-tokyo.ac.jp,
miyoshi@narr.jp,
hiroshi\_yamakawa@dwango.co.jp
}

\begin{document}

\maketitle

\begin{abstract}
We propose the factorized action variational autoencoder (FAVAE), a state-of-the-art generative model for learning disentangled and interpretable representations from sequential data via the information bottleneck without supervision. The purpose of disentangled representation learning is to obtain interpretable and transferable representations from data. We focused on the disentangled representation of sequential data since there is a wide range of potential applications if disentanglement representation is extended to sequential data such as video, speech, and stock market. Sequential data are characterized by dynamic and static factors: dynamic factors are time dependent, and static factors are independent of time. Previous models disentangle static and dynamic factors by explicitly modeling the priors of latent variables to distinguish between these factors. However, these models cannot disentangle representations between dynamic factors, such as disentangling "picking up" and "throwing" in robotic tasks. FAVAE can disentangle multiple dynamic factors. Since it does not require modeling priors, it can disentangle "between" dynamic factors. We conducted experiments to show that FAVAE can extract disentangled dynamic factors.
\end{abstract}

\section{Introduction}
\begin{figure}[t]
\centering
\begin{minipage}{0.3\hsize}
\includegraphics[width=\linewidth]{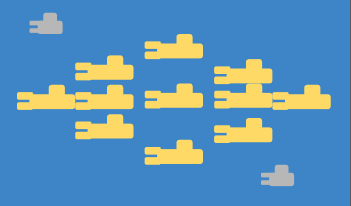}
\subcaption{$\beta$-VAE accepts sequential data}
\label{data_vae}
\end{minipage}
\begin{minipage}{0.68\hsize}
\includegraphics[width=\linewidth]{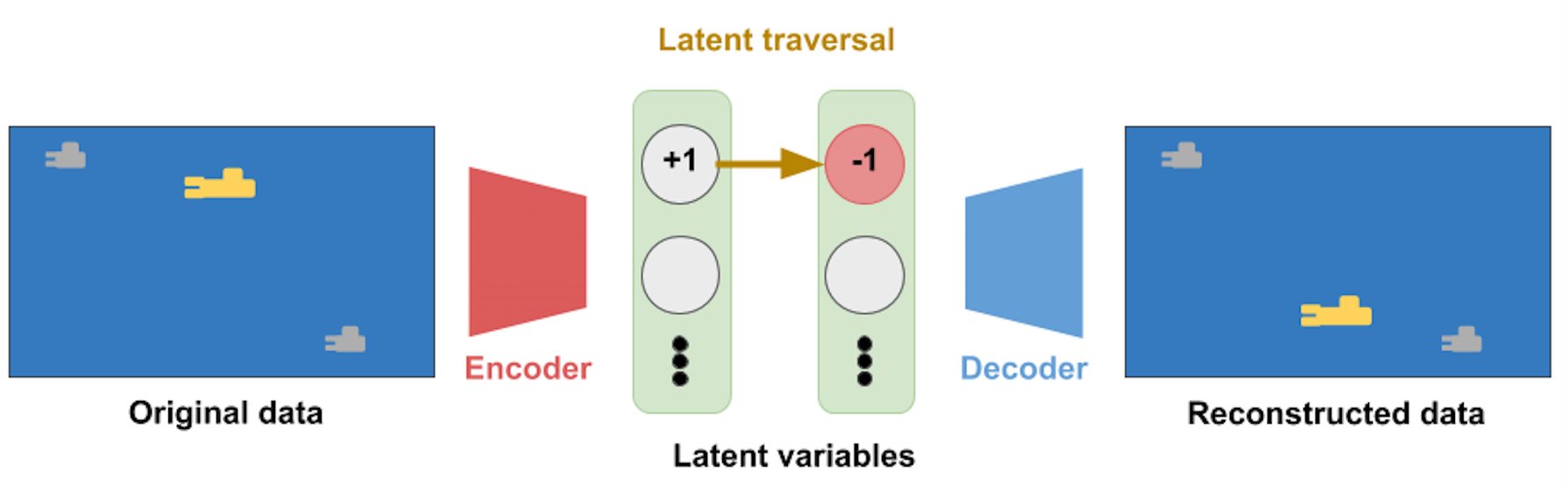}
\subcaption{Latent traversal of $\beta$-VAE}
\label{vae}
\end{minipage}
\begin{minipage}{0.3\hsize}
\includegraphics[width=\linewidth]{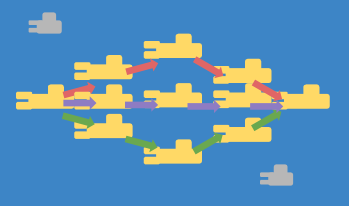}
\subcaption{FAVAE accepts sequential data}
\label{data_favae}
\end{minipage}
\begin{minipage}{0.68\hsize}
\includegraphics[width=\linewidth]{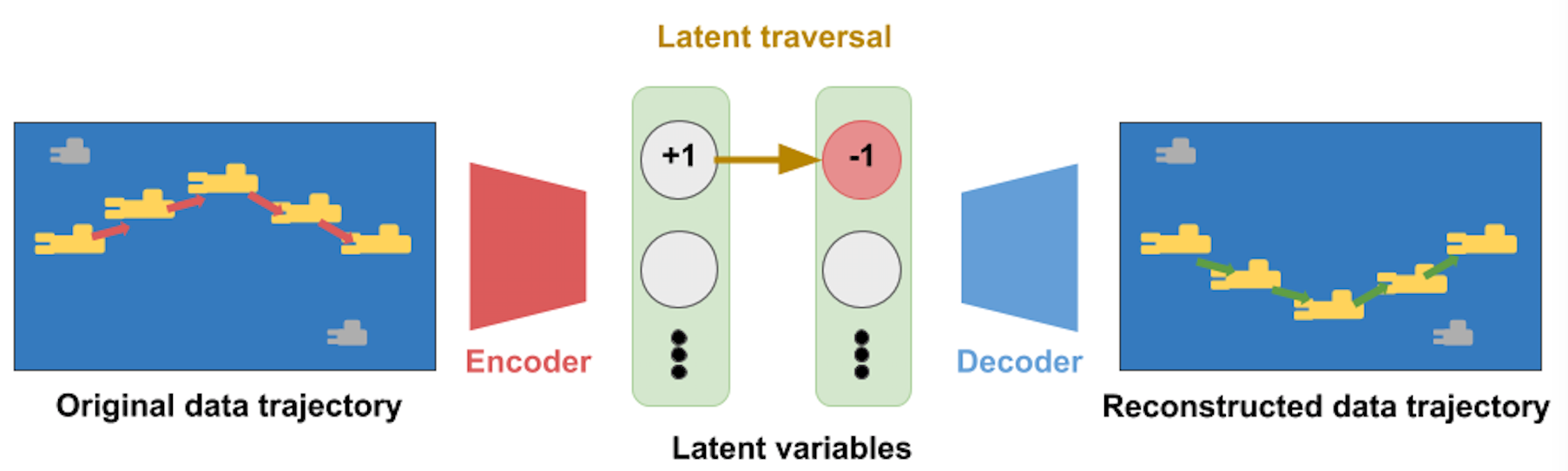}
\subcaption{Latent traversal of FAVAE}
\label{favae}
\end{minipage}
\caption{Illustration of how FAVAE differs from $\beta$-VAE, which does not accept data sequentially; it cannot differentiate data points from different trajectories or sequences of data points. FAVAE takes into account sequence of data points, taking all data points in trajectory as one datum. For example, for pseudo-dataset representing trajectory of submarine (\ref{data_vae},\ref{data_favae}), $\beta$-VAE accepts 11 different positions of submarine as non-sequential data, while FAVAE accepts three different trajectories of submarine as sequential data. Therefore, latent variable in $\beta$-VAE learns only coordinates of submarine, and latent traversal shows change in submarine’s position. However, FAVAE learns factor that controls trajectory of submarine, so latent traversal shows change in submarine’s trajectory.}
\label{fig:concept}
\end{figure}
Representation learning is one of the most fundamental problems in machine learning. A real-world data distribution can be regarded as a low-dimensional manifold in a high-dimensional space~\cite{Bengio:2013:RLR:2498740.2498889}. Generative models in deep learning, such as the variational autoencoder (VAE) ~\cite{vae} and generative adversarial network (GAN) ~\cite{gan}, can learn a low-dimensional manifold representation (factor) as a latent variable. The factors are fundamental components such as position, color, and degree of smiling in an image of a human face ~\cite{celeba}. Disentangled representation is defined as a single factor being represented by a single latent variable~\cite{Bengio:2013:RLR:2498740.2498889}. Thus, if in a model of learned disentangled representation, shifting one latent variable while leaving the others fixed generates data showing that only the corresponding factor was changed. This is called \textit{latent traversals} (a good demonstration of which was given by~\cite{beta-vae1}\footnote{This demonstration is available at \href{https://docs.google.com/presentation/d/12uZQ_Vbvt3tzQYhWR3BexqOzbZ-8AeT_jZjuuYjPJiY/pub?start=true&loop=true&delayms=30000&slide=id.g1329951dde_0_0}{http://tinyurl.com/jgbyzke}}). There are two advantages of disentangled representation. First, latent variables are interpretable. Second, the disentangled representation is generalizable and robust against adversarial attacks ~\cite{deepvariationalinformationbottleneck}.

We focus on the disentangled representation learning of sequential data. Sequential data are characterized by dynamic and static factors: dynamic factors are time dependent, and static factors are independent of time. With disentangled representation learning from sequential data, we should be able to extract dynamic factors that cannot be extracted using disentangled-representation-learning models, such as $\beta$-VAE ~\cite{beta-vae1,beta-vae2} and InfoGAN ~\cite{infogan}, for non-sequential data. The concept of disentangled representation learning for sequential data is illustrated in Fig. \ref{fig:concept}. Consider that the pseudo-dataset of the movement of a submarine has a dynamic factor: the trajectory shape. The disentangled-representation-learning model for sequential data can extract this shape. On the other hand, since the disentangled representation learning model for non-sequential data does not take into account the sequence of data, it merely extracts the x- and y-positions.

There is a wide range of potential applications if we extend disentanglement representation to sequential data such as speech, video, and stock market. For example, disentangled representation learning for stock-market data can extract the fundamental trend of a given stock price. Another application is the reduction of action space in reinforcement learning. Extracting dynamic factors would enable the generation of macro-actions~\cite{dqn-macro-action}, which are sets of sequential actions that represent the fundamental factors of the actions. Thus, disentangled representation learning for sequential data opens the door to new areas of research.

Very recent related work~\cite{fhvae,disney} separated factors of sequential data into dynamic and static factors. The factorized hierarchical VAE (FHVAE) ~\cite{fhvae} is based on a graphical model using latent variables with different time dependencies. By maximizing the variational lower bound of the graphical model, FHVAE separates the different time-dependent factors, such as dynamic, from static factors. The VAE architecture developed by~\cite{disney} is the same as that in the FHVAE in terms of the time dependencies of the latent variables. Since these models require different time dependencies for the latent variables, they cannot be used to disentangle variables with the same time-dependency factor.

We address this problem by taking a different approach. First, we analyze the root cause of disentanglement from the perspective of information theory. As a result, the term causing disentanglement is derived from a more fundamental rule: reduce the mutual dependence between the input and output of an encoder while keeping the reconstruction of the data. This is called the information bottleneck (IB) principle. We naturally extend this principle to sequential data from the relationship between $x$ and $z$ to $x_{t:T}$ and $z$. This enables the separation of multiple dynamic factors as a consequence of information compression. It is difficult to learn a disentangled representation of sequential data since not only the feature space but also the time space should be compressed. We developed the factorized action VAE (FAVAE) in which we implemented the concept of information capacity to stabilize learning and a ladder network to learn a disentangled representation in accordance with the level of data abstraction. Since FAVAE is a more general model without the restriction of a graphical model design to distinguish between static and dynamic factors, it can separate dependency factors occurring at the same time. It can also separate factors into dynamic and static.

\section{Disentanglement for Non-Sequential Data}
The $\beta$-VAE ~\cite{beta-vae1,beta-vae2} is commonly used for learning disentangled representations based on the VAE framework~\cite{vae} for a generative model. The VAE can estimate the probability density from data x. The objective function of the VAE maximizes the evidence lower bound (ELBO) of $\log p\left(x\right)$ as 
\begin{align}
\log p\left(x\right) & =\mathcal{L}_{{\rm VAE}}+\underbrace{D_{{\rm KL}}\left(q\left(z|x\right)||p\left(z|x\right)\right)}_{\geq0},
\end{align}
where $z$ is a latent variable, $D_{{\rm KL}}$ is the Kullback-Leibler divergence, and $q\left(z|x\right)$ is an approximated distribution of $p\left(z|x\right)$. The $D_{{\rm KL}}\left(q\left(z|x\right)||p\left(z|x\right)\right)$ reduces to zero as the ELBO $\mathcal{L}_{{\rm VAE}}$ increases; thus, $q\left(z|x\right)$ learns a good approximation of $p\left(z|x\right)$. The ELBO is defined as 
\begin{equation}
\mathcal{L}_{{\rm VAE}}\equiv E_{q\left(z|x\right)}\left[\log p\left(x|z\right)\right]-D_{{\rm KL}}\left(q\left(z|x\right)||p\left(z\right)\right),
\end{equation}
where the first term $E_{q\left(z|x\right)}\left[\log p\left(x|z\right)\right]$ is a reconstruction term used to reconstruct $x$, and the second term $D_{{\rm KL}}\left(q\left(z|x\right)||p\left(z\right)\right)$ is a regularization term used to regularize posterior $q\left(z|x\right)$. Encoder $q\left(z|x\right)$ and decoder $p\left(x|z\right)$ are learned in the VAE. 

Next, we explain how $\beta$-VAE extracts disentangled representations from unlabeled data. The $\beta$-VAE is an extension of the coefficient $\beta>1$ of $D_{{\rm KL}}\left(q\left(z|x\right)||p\left(z\right)\right)$ in the VAE. The objective function of $\beta$-VAE is
\begin{equation}
\mathcal{L}_{{\rm \beta-VAE}}=E_{q\left(z|x\right)}\left[\log p\left(x|z\right)\right]-\beta D_{{\rm KL}}\left(q\left(z|x\right)||p\left(z\right)\right),\label{eq:betavea}
\end{equation}
where $\beta>1$ and $p\left(z\right)=\mathcal{N}\left(0,1\right)$. The $\beta$-VAE promotes disentangled representation learning via $D_{{\rm KL}}\left(q\left(z|x\right)||p\left(z\right)\right)$. As $\beta$ increases, the latent variable $q\left(z|x\right)$ approaches the prior $p\left(z\right)$; therefore, each $z_{i}$ is forced to learn the probability distribution of $\mathcal{N}\left(0,1\right)$. However, if all latent variables $z_{i}$ become $\mathcal{N}\left(0,1\right)$, the model cannot reconstruct $x$. As a result, as long as $z$ reconstructs $x$, $\beta$-VAE reduces the information of $z$.

\section{Preliminary: Origin of Disentanglement\label{subsec:Origin-of-Disentanglement}}
To clarify the origin of disentanglement, we explain the regularization term. The regularization term has been decomposed into three terms ~\cite{isolating,factorvae,matthew}: 
\begin{align}
E_{p\left(x\right)}\left[D_{{\rm KL}}\left(q\left(z|x\right)||p\left(z\right)\right)\right]&=I\left(x;z\right)\nonumber\\
+D_{{\rm KL}}\left(q\left(z\right)||\prod_{j}q\left(z_{j}\right)\right)&+\sum_{j}D_{{\rm KL}}\left(q\left(z_{j}\right)||p\left(z_{j}\right)\right),\label{eq:reg}
\end{align}
where $z_{j}$ denotes the $j$-th dimension of the latent variable. The second term, which is called "total correlation" in information theory, quantifies the redundancy or dependency among a set of n random variables ~\cite{totalorrelation}. The $\beta$-TCVAE ~\cite{isolating} has been experimentally shown to reduce the total correlation causing disentanglement. The third term indirectly causes disentanglement by bringing $q\left(z|x\right)$ close to the independent standard normal distribution $p\left(z\right)$. The first term is mutual information between the data variable and latent variable based on the empirical data distribution. Minimizing the regularization term causes disentanglement but disrupts reconstruction via the first term in Eq. (\ref{eq:reg}). The shift $C$ scheme was proposed~\cite{understand-beta-vae} to solve this conflict:
\begin{equation}
-E_{q_{\phi}\left(z|x\right)}\left[\log p_{\theta}\left(x|z\right)\right]+\beta\left|D_{{\rm KL}}\left(q\left(z|x\right)||p\left(z\right)\right)-C\right|,\label{eq:bcvae}
\end{equation}
where constant shift $C$, which is called "information capacity," linearly increases during training. This $C$ can be understood from the point of view of an information bottleneck~\cite{informationbottleneck}. The VAE can be derived by maximizing the ELBO, but $\beta$-VAE can no longer be interpreted as an ELBO once this scheme has been applied. The objective function of $\beta$-VAE is derived from the information bottleneck ~\cite{deepvariationalinformationbottleneck,informationdropout,informationbottleneck,informationbottleneckgaussinan}.
\begin{equation}
\max I\left(z;x\right)\qquad{\rm s.t.}\:\left|I\left(\hat{x};z\right)-C\right|=0,\label{eq:ib}
\end{equation}
where $\hat{x}$ is the empirical distribution. Solving this equation by using Lagrange multipliers drives the objective function of $\beta$-VAE (Eq. (\ref{eq:bcvae})) with $\beta$ as the Lagrange multiplier (details in Appendix B of~\cite{deepvariationalinformationbottleneck}). In Eq. (\ref{eq:bcvae}),  $C$ prevents $I\left(\hat{x},z\right)$ from becoming zero. In the literature on information bottleneck, $y$ typically stands for a classification task; however, the formulation can be related to the autoencoding objective~\cite{deepvariationalinformationbottleneck}. Therefore, the objective function of $\beta$-VAE can be understood using the IB principle.

\section{Proposed Model: Disentanglement for Sequential Data}
FAVAE learns disentangled and interpretable representations from sequential data without supervision. We consider sequential data $x_{1:T}\equiv\left\{ x_{1},x_{2},\cdots,x_{T}\right\} $ generated from a latent variable model, 
\begin{align}
p\left(x_{1:T}\right) & =\int p\left(x_{1:T}|z\right)p\left(z\right)dz.
\end{align}
For sequential data, we replace $x$ with $\left(x_{1:T}\right)$ in Eq. \ref{eq:bcvae}. The objective function of FAVAE is
\begin{align}
&-E_{q_{\phi}\left(z|\left(x_{1:T}\right)_{I}\right)}\left[\log p_{\theta}\left(x_{1:T}|z\right)\right]\nonumber\\
&+\beta\left|D_{{\rm KL}}\left(q\left(z|\left(x_{1:T}\right)_{I}\right)||p\left(z\right)\right)-C\right|,\label{eq:tcvlae_loss}
\end{align}
where $p\left(z\right)=\mathcal{N}\left(0,1\right)$. The variational recurrent neural network ~\cite{VRNN} and stochastic recurrent neural network (SRNN) ~\cite{SRNN} extend the VAE to a recurrent framework. The priors of both networks are dependent on time. The time-dependent prior experimentally improves the ELBO. In contrast, the prior of FAVAE is independent of time like those of the stochastic recurrent network (STORN)~\cite{STORN} and Deep Recurrent Attentive Writer (DRAW) neural network architecture~\cite{DRAW}; this is because FAVAE is disentangled representation learning rather than density estimation. For better understanding, consider FAVAE from the perspective of information bottleneck. As with $\beta$-VAE, FAVAE can be understood from the IB principle.
\begin{equation}
\max I\left(z;x_{1:T}\right)\qquad{\rm s.t.}\:\left|I\left(\hat{x}_{1:T};z\right)-C\right|=0,
\end{equation}
where $\hat{x}_{1:T}$ follows an empirical distribution. These principles make the representation of $z$ compact, while reconstruction of the sequential data is represented by $x_{1:T}$ (see Appendix A).

\subsection{Ladder Network}
\begin{figure}
 \includegraphics[width=0.5\textwidth]{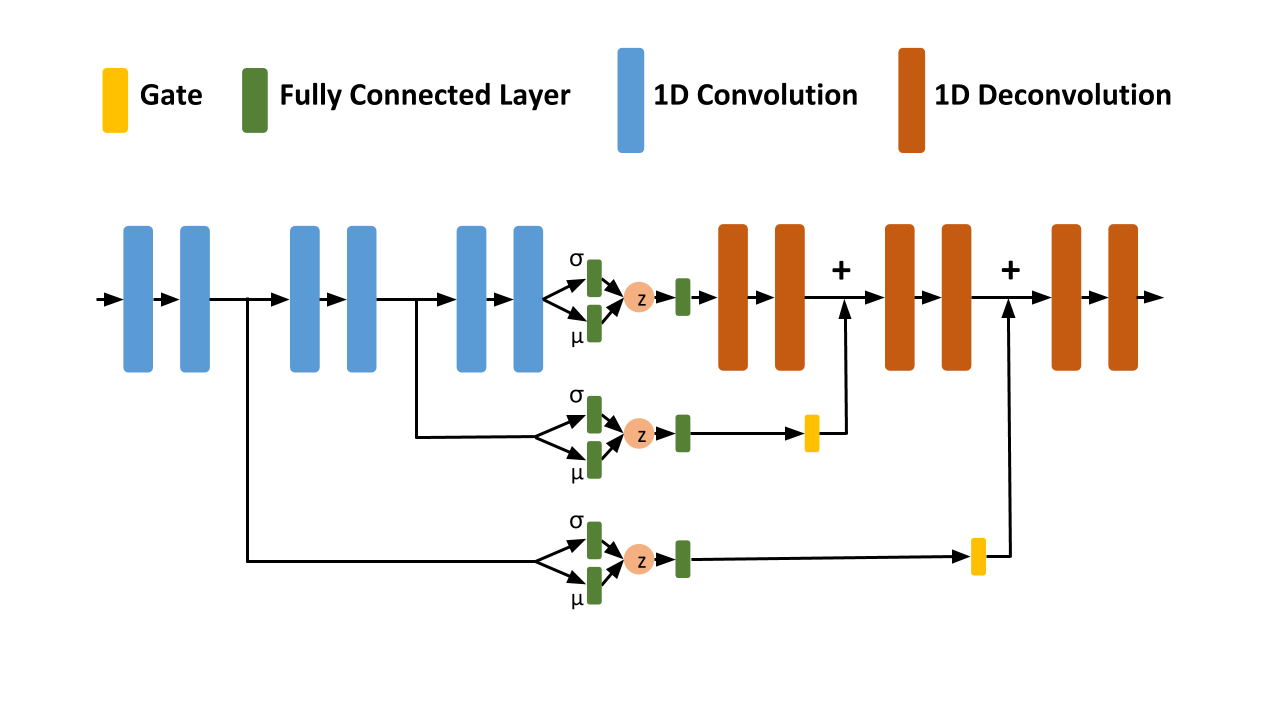}
\caption{FAVAE architecture}
\label{fig:Architecture-of-TCVLAE.}
\end{figure}

An important extension to FAVAE is a hierarchical representation scheme inspired by the variational ladder AE (VLAE)~\cite{vlae}. Encoder $q\left(z|x_{1:T}\right)$ within a ladder network is defined as 
\begin{align}
h_{l} & =f_{l}\left(h_{l-1}\right),\\
z_{l} & \sim\mathcal{N}\left(\mu_{l}\left(h_{l}\right),\sigma_{l}\left(h_{l}\right)\right),
\end{align}
where $l$ is a layer index, $h_{0}\equiv x_{1:T}$, and $f$ is a time-convolution network, which is explained in the next section. Decoder $p\left(x_{1:T}|z\right)$ within the ladder network is defined as 
\begin{align}
\tilde{z}_{l} & =g_{l}\left(z_{l}\right)\\
\tilde{z}_{l} & =g_{l}\left(\tilde{z}_{l+1}+{\rm gate}\left(z_{l}\right)\right),\\
x_{1:T} & \sim r\left(x_{1:T},\tilde{z}_{0}\right)
\end{align}
where $g_{l}$ is the time deconvolution network with $l=1,\cdots,L-1$, and $r$ is a distribution family parameterized by $g_{0}\left(\tilde{z}_{0}\right)$. The gate computes the Hadamard product of its learnable parameter and input tensor. We set $r$ as a fixed-variance factored Gaussian distribution with the mean given by $\mu_{t:T}=g_{0}\left(\tilde{z}_{0}\right)$. Figure (\ref{fig:Architecture-of-TCVLAE.}) shows the architecture of FAVAE. The difference between each ladder network in the model is the number of convolution networks through which data passes. The abstract expressions should differ between ladders since the time-convolution layer abstracts sequential data. Without the ladder network, FAVAE can disentangle only the representations at the same level of abstraction; with the ladder network, it can disentangle representations at different levels of abstraction.

\subsection{How to encode and decode}
There are several mainstream neural network models designed for sequential data, such as the long short-term memory (LSTM) model~\cite{lstm}, gated recurrent unit model~\cite{gru}, and quasi-recurrent neural network QRNN~\cite{qrnn}. However, VLAE has a hierarchical structure created by abstracting a convolutional neural network, so it is simple to add the time convolution of the QRNN to FAVAE. The input data are $x_{t,i}$, where $t$ is the time index and $i$ is the dimension of the feature vector index. The time convolution takes into account the dimensions of feature vector $j$ as a convolution channel and performs convolution in the time direction: 
\begin{equation}
z_{tj}=\sum_{p,i} x_{t-p,i}h_{pij}+b_{j},
\end{equation}
where $j$ is the channel index. FAVAE has a network similar to that of VAE regarding time convolution and a loss function similar to that of $\beta$-VAE (Eq. (\ref{eq:tcvlae_loss})). We use the batch normalization~\cite{batchnorm} and rectified linear unit as activation functions, though other variations are possible. For example, 1D convolutional neural networks use a filter size of 3 and stride of 2 and do not use a pooling layer.

\begin{figure}
\centering
\begin{minipage}{0.17\hsize}
\includegraphics[width=\linewidth]{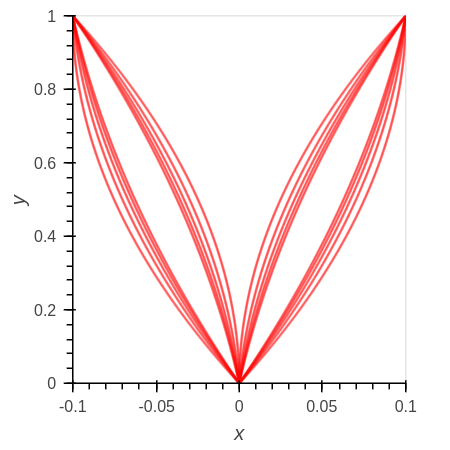}
\subcaption{}
\label{2d_reaching_traindata}
\end{minipage}
\begin{minipage}{0.17\hsize}
\includegraphics[width=\linewidth]{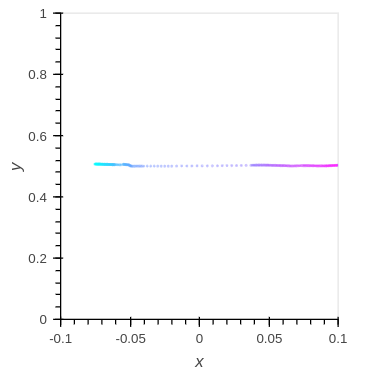}
\subcaption{}
\label{beta_vae_z0_to_x}
\end{minipage}
\begin{minipage}{0.17\hsize}
\includegraphics[width=\linewidth]{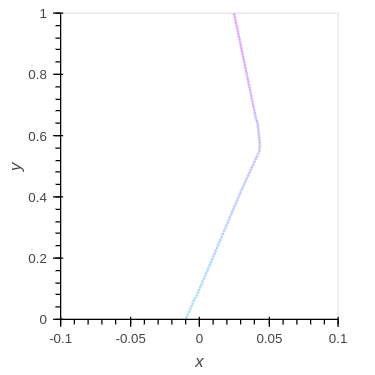}
\subcaption{}
\label{beta_vae_z1_to_x}
\end{minipage}
\begin{minipage}{0.17\hsize}
\includegraphics[width=\linewidth]{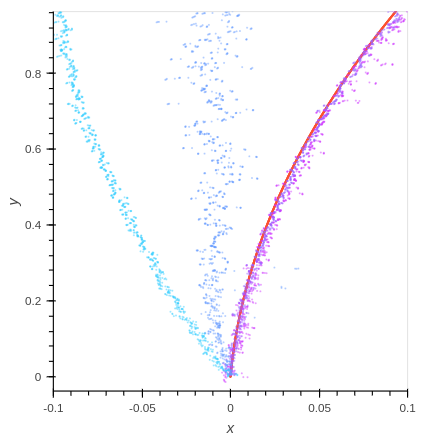}
\subcaption{}
\label{favae_z1_to_x}
\end{minipage}
\begin{minipage}{0.17\hsize}
\includegraphics[width=\linewidth]{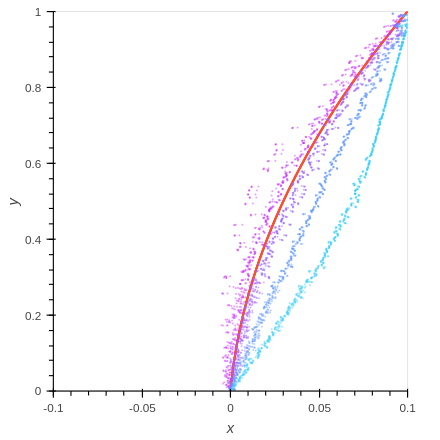}
\subcaption{}
\label{favae_z2_to_x}
\end{minipage}
\caption{Visualization of latent traversal of $\beta$-VAE and FAVAE. \ref{2d_reaching_traindata} represents all data trajectories of 2D reaching. \ref{beta_vae_z0_to_x} and \ref{beta_vae_z1_to_x} represent latent traversal in $\beta$-VAE,  \ref{favae_z1_to_x} and \ref{favae_z2_to_x} represent latent traversal in FAVAE. Each latent variable is traversed and purple and/or blue points are generated. The color corresponds to the value of the traversed latent variable.}
\label{fig:2dreach}
\end{figure}

\begin{figure}[t]
\includegraphics[width=\linewidth]{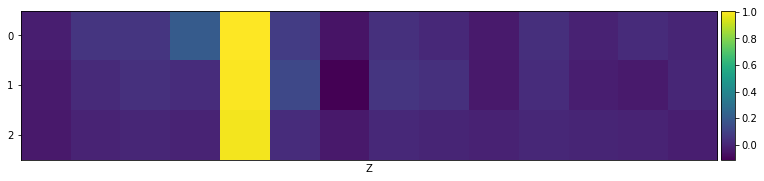}
\centering
\caption{Visualization of $\frac { I \left( z _ { j } ; v _ { k } \right) } { H \left( v _ { k } \right) }$. Horizontal axis shows latent variable and vertical axis shows factor. It shows case in which all information concentrates in 4th latent variable in 2D Reaching.}
\label{fig:bad_mig_table}
\end{figure}

\section{Measuring Disentanglement}
While latent traversals are useful for checking the success or failure of disentanglement, quantification of the disentanglement is required for reliably evaluating a learned model. Various disentanglement quantification methods have been reported~\cite{eastwood2018a,isolating,factorvae,beta-vae2,beta-vae1}, but there is no standard method. We use the mutual information gap (MIG)~\cite{isolating} as the metric for disentanglement. The basic idea of MIG is measuring the mutual information between latent variables $z_{j}$ and a ground-truth factor $v_{k}$. Higher mutual information means that $z_{j}$ contains more information regarding $v_{k}$. 
\begin{equation}
{\rm MIG}\equiv\frac{1}{K}\sum_{k=1}^{K}\frac{1}{H\left(v_{k}\right)}\left(I\left(z_{j^{\left(k\right)}};v_{k}\right)-\max_{j\neq j^{\left(k\right)}}I\left(z_{j};v_{k}\right)\right),
\end{equation}
where $j^{\left(k\right)}\equiv\arg\max_{j}I\left(z_{j};v_{k}\right)$, and $H\left(v_{k}\right)$ is entropy for normalization.

There is a problem with MIG when measuring with simple data. When it is possible to reconstruct with one latent variable, using large $\beta$ gathers all factor information in one latent variable and MIG becomes large (Fig. \ref{fig:bad_mig_table}). For example, when goal position, curved inward/outward, and degree of curvature cannot be disentangled to different latent variables in the 2D Reaching dataset, MIG can become large. In our experiments we avoided this problem by excluding the case in which all factor information concentrates in one latent variable.

\section{Related Work}
Several recently reported models~\cite{fhvae,disney} graphically disentangle static and dynamic factors in sequential data such as speech and video~\cite{timit,aurora}. These models learn by building the time dependency of the prior of the latent variable. In particular, FHVAE~\cite{fhvae} uses label data that distinguish time series for learning. Note that the label is not a dynamic factor but a label to distinguish between time series. In contrast, FAVAE performs disentanglement by using a loss function (see Eq. \ref{eq:tcvlae_loss}). The advantage of graphical models is that they can control the interpretable factors by controlling the prior’s time dependency. Since dynamic factors have the same time dependency, these models cannot disentangle dynamic factors. Since FAVAE has no time-dependency constraint of the prior, it can disentangle static and dynamic factors as well as disentangle sets of dynamic factors.

\section{Experiments}
We experimentally evaluated FAVAE using five sequential datasets: 2D Reaching with sequences 100 and 1000, 2D Wavy Reaching with sequences 100 and 1000, and Sprites dataset~\cite{disney}. We used a batch size of $128$ and the Adam~\cite{adam} optimizer with a learning rate of $10^{-3}$.

\subsection{2D Reaching}
To determine the differences between FAVAE and $\beta$-VAE, we used a bi-dimensional space reaching dataset. Starting from point (0, 0), the point travels to goal position (-0.1, +1) or (+0.1, +1). There are ten possible trajectories to each goal; five are curved inward, and the other five are curved outward. The degree of curvature for all five trajectories is different. The number of factor combinations was thus 20 (2x2x5). The trajectory lengths were 100 and 1000.

We compared the performances of $\beta$-VAE and FAVAE trained on the 2D Reaching dataset. The results of latent traversal are transforming one dimension of latent variable z into another value and reconstructing the output data from the traversed latent variables. The $\beta$-VAE, which is only able to learn from every point of a trajectory separately, encodes data points into latent variables that are parallel to the x and y axes (\ref{beta_vae_z0_to_x}, \ref{beta_vae_z1_to_x}). In contrast, FAVAE learns through one entire trajectory and can encode disentangled representations effectively so that feasible trajectories are generated from traversed latent variables (\ref{favae_z1_to_x}, \ref{favae_z2_to_x}).

\subsection{2D Wavy Reaching}

\begin{table*}[t]
\centering
\caption{Disentanglement scores (MIG and reconstruction loss) with standard deviations by repeating experiment 10 times for different models. Best results are shown in bold. ($L)$ means with ladder and ($C$) means with information capacity (e.g. FAVAE (L-) means FAVAE with ladder network without information capacity).}
\label{tab:mig_rec}
\begin{tabular}{lllllllll}
\toprule
\multirow{3}{*}{Model} & \multicolumn{4}{c}{2D Reaching} & \multicolumn{4}{c}{2D Wavy Reaching}\\ \cmidrule(l){2-9}
& \multicolumn{2}{c}{length=100} & \multicolumn{2}{c}{length=1000}& \multicolumn{2}{c}{length=100} & \multicolumn{2}{c}{length=1000} \\ \cmidrule(l){2-9} 
 & MIG & Rec & MIG & Rec & MIG & Rec & MIG & Rec \\ \midrule
FHVAE & {\bf 0.43(14)} & {\bf0.0013(23) } & - & - & 0.22(8) & 0.043(61) & - & - \\
FAVAE (L) ($\beta=0$) & 0.06(3) & 0.022(22) & 0.05(4) & 0.493(790) & 0.02(1) & {\bf 0.015(5)} & 0.04(3) & {\bf 0.085(17)} \\
FAVAE (- -) & 0.07(12) & 0.257(173) & 0.46(18) & 2.209(1869) & 0.66(15) & 0.041(8) & {\bf 0.47(18)} & 11.881(24014) \\
FAVAE (- C) & 0.09(13) & 0.257(172) & 0.46(18) & 1.193(1274) & {\bf 0.67(16)} & 0.042(21) & 0.31(10) & 5.937(18033) \\
FAVEA (L -) & 0.28(21) & 0.006(4) & 0.43(6) & 0.022(9) & 0.29(9) & 0.123(16) & 0.28(4) & 0.707(86) \\
FAVAE (L C) & 0.28(11) & 0.008(14) & {\bf 0.64(6)} & {\bf 0.017(6)} & 0.42(17) & 0.046(11) & 0.24(7) & 0.190(95) \\ \bottomrule
\end{tabular}
\end{table*}

To confirm the effect of disentanglement through the information bottleneck, we evaluated the validity of FAVAE under more complex factors by adding more factors to 2D Reaching. Five factors in total generated data compared to the three factors that generate data in 2D Reaching. This modified dataset differed in that four out of the five factors affect only part of the trajectory: two affected the first half, and the other two affected the second half. This means that the model should be able to focus on a certain part of the whole trajectory and extract factors related to that part. A detailed explanation of these factors is given in Github \footnote{Dataset is available at \href{https://github.com/favae/favae_ijcai2019}{https://github.com/favae/favae\_ijcai2019}}.

\begin{figure}[htpb]
\centering
\begin{minipage}{0.24\hsize}
\includegraphics[width=\linewidth]{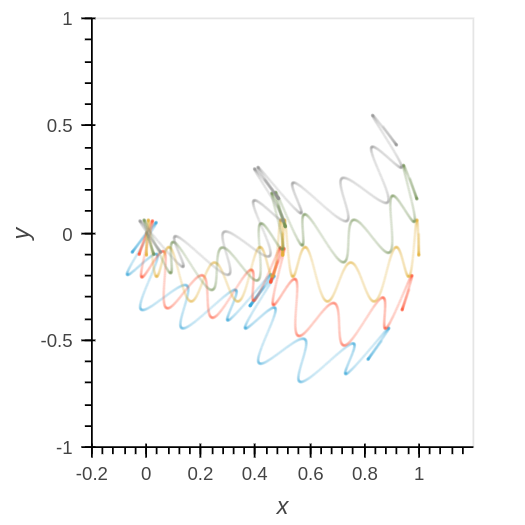}
\subcaption{factor 1}
\label{fig:2dwavy_3}
\end{minipage}
\begin{minipage}{0.24\hsize}
\includegraphics[width=\linewidth]{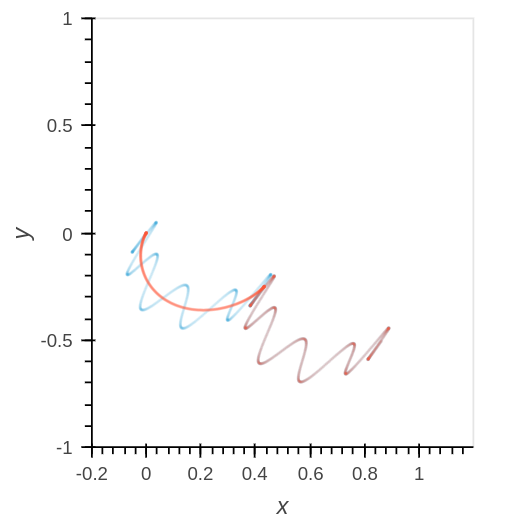}
\subcaption{factor 2}
\label{fig:2dwavy_1}
\end{minipage}
\begin{minipage}{0.24\hsize}
\includegraphics[width=\linewidth]{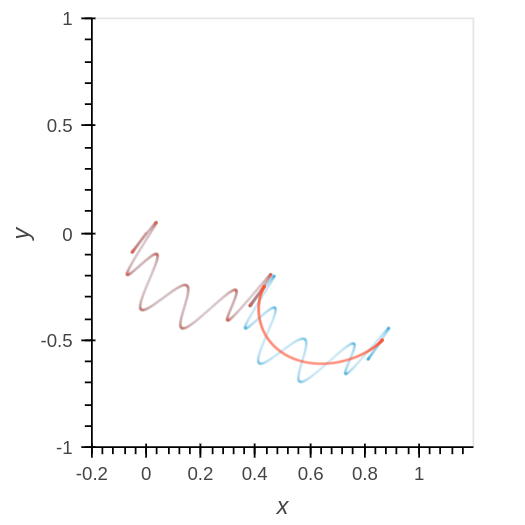}
\subcaption{factor 3}
\label{fig:2dwavy_2}
\end{minipage}
\begin{minipage}{0.24\hsize}
\includegraphics[width=\linewidth]{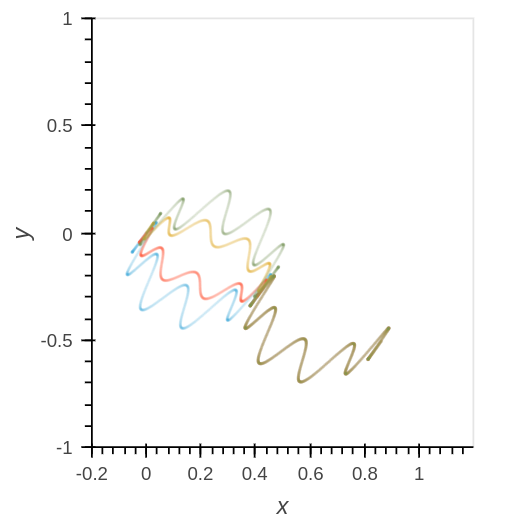}
\subcaption{factor 4}
\label{fig:2dwavy_4}
\end{minipage}
\begin{minipage}{0.24\hsize}
\includegraphics[width=\linewidth]{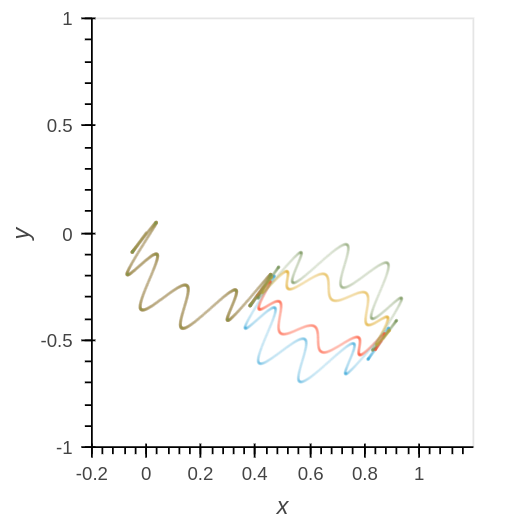}
\subcaption{factor 5}
\label{fig:2dwavy_5}
\end{minipage}%
\begin{minipage}{0.24\hsize}
\includegraphics[width=\linewidth]{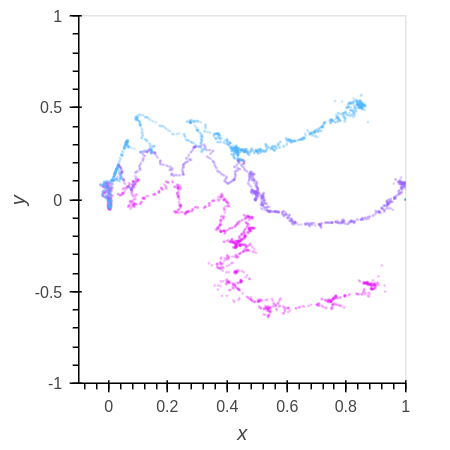}
\subcaption{factor 1}
\label{fig:2dwavy_latent_traversal_1}
\end{minipage}
\begin{minipage}{0.24\hsize}
\includegraphics[width=\linewidth]{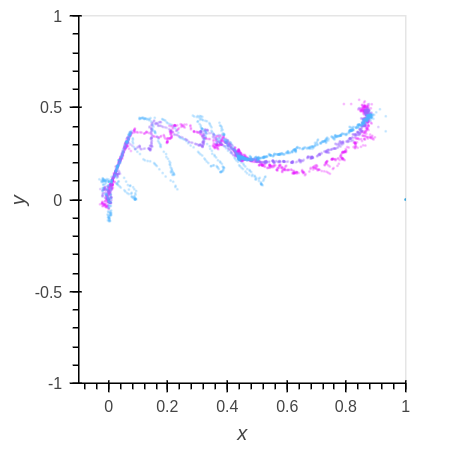}
\subcaption{factor 2}
\label{fig:2dwavy_latent_traversal_2}
\end{minipage}
\begin{minipage}{0.24\hsize}
\includegraphics[width=\linewidth]{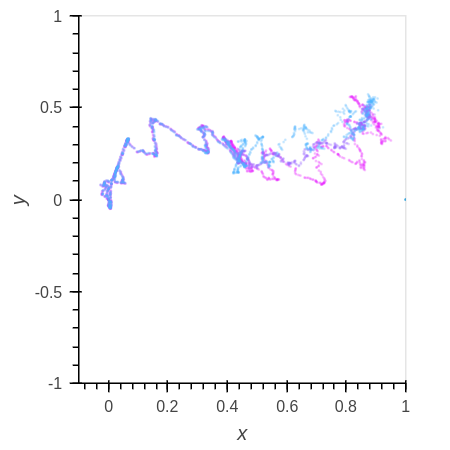}
\subcaption{factor 3}
\label{fig:2dwavy_latent_traversal_3}
\end{minipage}
\begin{minipage}{0.24\hsize}
\includegraphics[width=\linewidth]{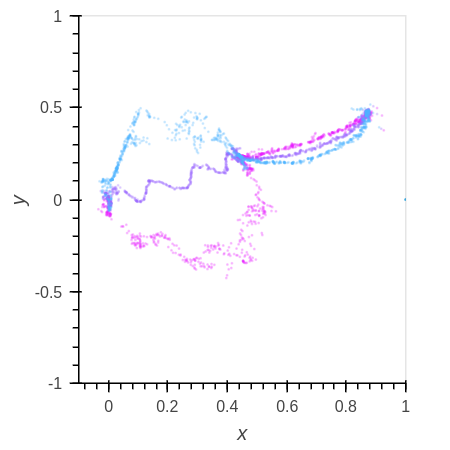}
\subcaption{factor 4}
\label{fig:2dwavy_latent_traversal_4}
\end{minipage}
\begin{minipage}{0.24\hsize}
\includegraphics[width=\linewidth]{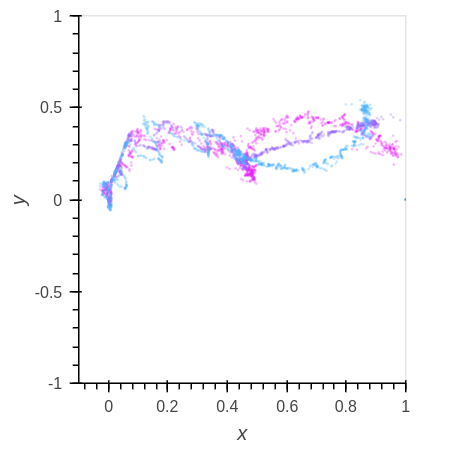}
\subcaption{factor 5}
\label{fig:2dwavy_latent_traversal_5}
\end{minipage}
\caption{Visualization of training data (\ref{fig:2dwavy_3} to \ref{fig:2dwavy_5}) and latent traversal (\ref{fig:2dwavy_latent_traversal_1} to \ref{fig:2dwavy_latent_traversal_5}) for 2D Wavy Reaching. The vertical and horizontal axes represent coordinates. Factors 1, 2, 3, 4, and 5 respectively correspond to "Goal position", "1st trajectory shape", "2nd trajectory shape", "1st trajectory degree of curvature" and "2nd trajectory degree of curvature". Each plot was decoded by traversing one latent variable; different colors represent trajectories generated from different values of same latent variable, z.}
\label{fig:2dwave_z2x}
\end{figure}

We show the training dataset of 2D Wavy Reaching and latent traversal in FAVAE (LC) with sequence length 1000 in Fig. \ref{fig:2dwave_z2x}. The latent traversal results for 2D Wavy Reaching are plotted in Figs. \ref{fig:2dwavy_latent_traversal_1} to \ref{fig:2dwavy_latent_traversal_5}. Even though not all learned representations were perfectly disentangled, the visualization shows that all five generation factors were learned from five latent variables; the other latent variables did not learn any meaningful factors, indicating that the factors could be expressed as a combination of five "active" latent variables.

We compared various models on the basis of MIG to demonstrate the validity of FAVAE, i.e., time convolution AE in which a loss function is used only for the AE ($\beta=0$), FAVAE with/without the ladder network ($L$) and information capacity ($C$), and FHVAE~\cite{fhvae} which is the recently proposed disentangled representation learning model, as the baseline. Note that FHVAE uses label information (this label for distinguishing time series is not a dynamic factor) to disentangle time series data, which is a different setup with FAVAE. Table \ref{tab:mig_rec} shows a comparison of MIG scores and reconstruction losses using FHVAE as the baseline for 2D Reaching and 2D Wavy Reaching each with sequence lengths of 100 and 1000. In 2D Reaching, the MIG of the baseline was large, while in 2D Wavy Reaching the MIG of FAVAE was large. This is because FHVAE uses goal-position information as a label when learning. Even when there were multiple dynamic factors such as in 2D Wavy Reaching, FAVAE exhibited good disentangle performance (the large MIG and the small reconstruction loss).

When the ladder was added, the reconstruction loss was stable (especially at sequence length 1000). For example, looking at the length = 1000 of 2D Wavy Reaching in Table \ref{tab:mig_rec}, without ladder had a large MIG but the distribution of reconstruction loss was very large.

\begin{figure}
\centering
\begin{minipage}{0.45\hsize}
\includegraphics[width=\linewidth]{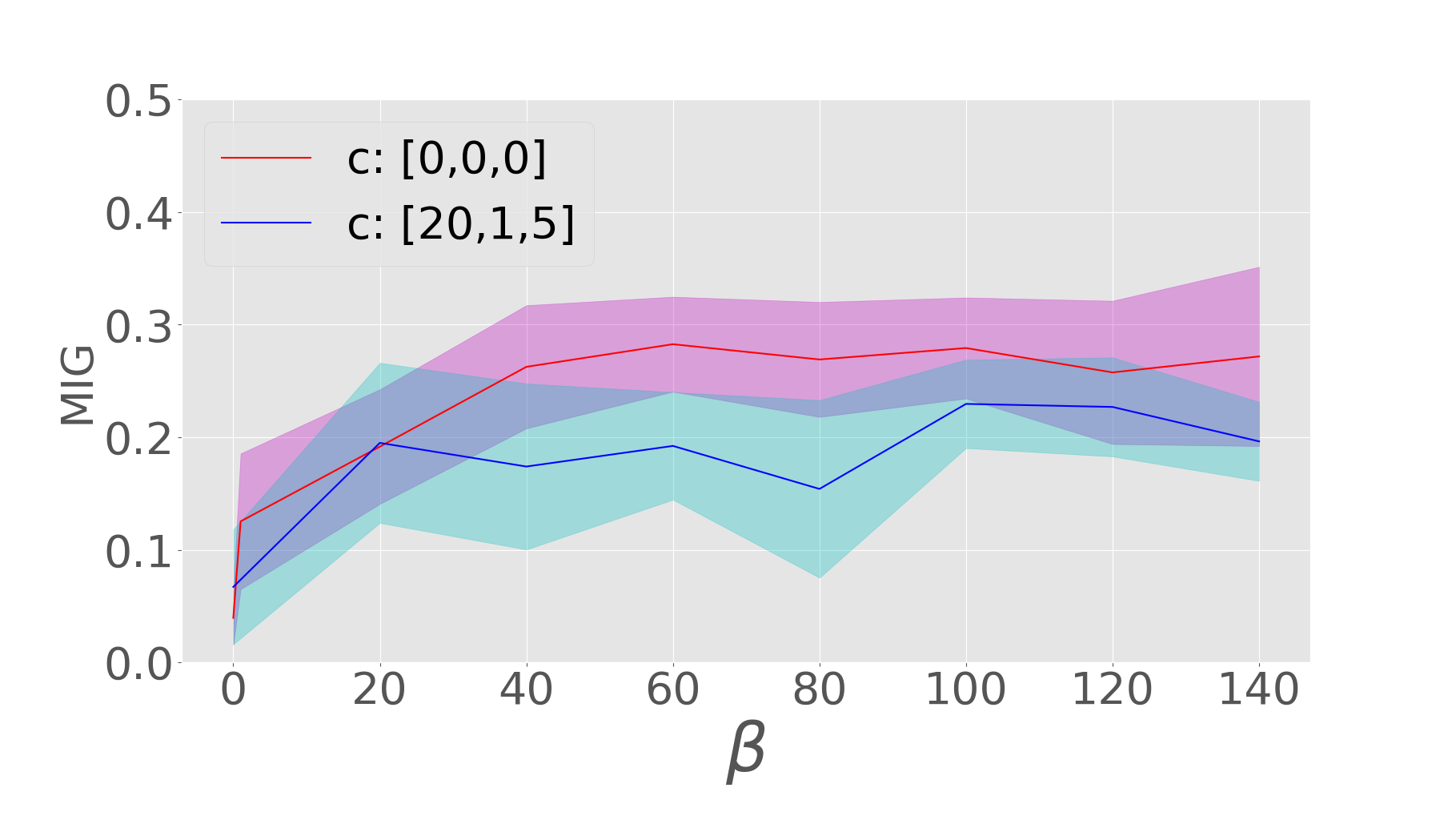}
\end{minipage}
\begin{minipage}{0.48\hsize}
\includegraphics[width=\linewidth]{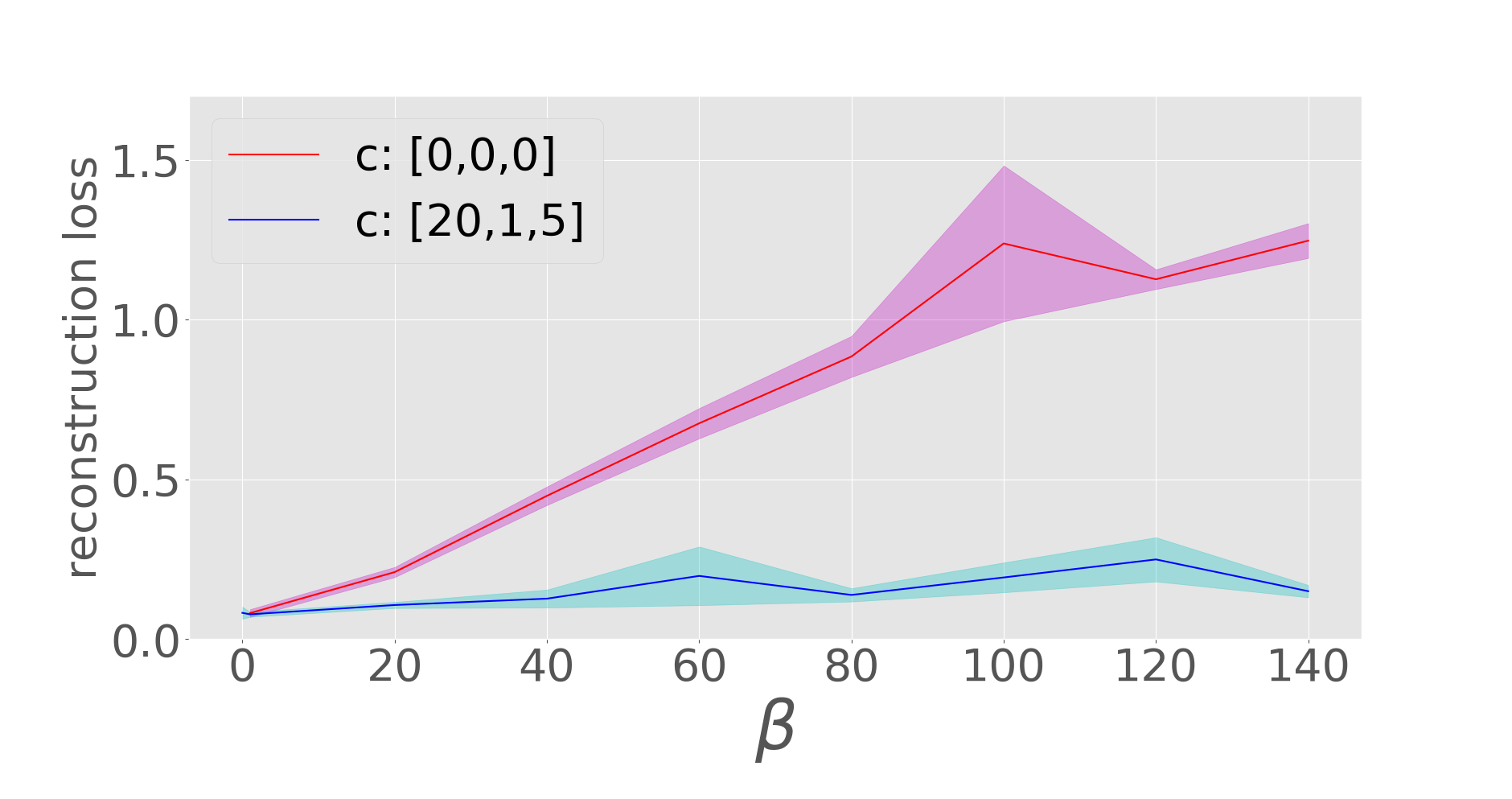}
\end{minipage}
\caption{MIG scores and reconstruction losses for different $\beta$. Blue line represents results with information capacity $C$ greater than zero; red line represents results with $C$ set to zero. Note that x axis is plotted in log scale.}
\label{fig:mig_entropy_rec}
\end{figure}

To confirm the effect of $C$, we the evaluated reconstruction losses and MIG scores for various $\beta$ using three ladder networks (Fig. \ref{fig:Architecture-of-TCVLAE.}) with a different $C$ for each ladder: $C=\left[\rm{LowerLadder},\rm{MiddleLadder},\rm{HigherLadder}\right]$ in Fig. \ref{fig:mig_entropy_rec}. One setting was $C=[0,0,0]$, meaning that $C$ was not used; another setting was $C=[20,1,5]$, meaning that $C$ was adjusted on the basis of KL-Divergence for $\beta=1$ and $C=[0,0,0]$. When $C$ was not used, FAVAE could not reconstruct data when $\beta$ was high; thus, disentangled representation was not learned well when $\beta$ was high. When $C$ was used, the MIG score increased with $\beta$ while reconstruction loss was suppressed.

\begin{table}[t]
\centering
\caption{For each factor, counting the number of latent variables which is the highest $I(v_k, z)$ in each ladder (1st, 2nd, 3rd). The same operation is performed ten times and results are shown. The detail of factor is shown in Github\protect\footnotemark[2]}
\label{tab:ladder_level}
\begin{tabular}{@{}llccc@{}}
\toprule
 &  & \multicolumn{1}{l}{1st} & \multicolumn{1}{l}{2nd} & \multicolumn{1}{l}{3rd} \\ \midrule
\multicolumn{1}{c}{\multirow{3}{*}{2D Reaching}} & \multicolumn{1}{l|}{factor 1} & 1 & 1 & {\bf 8} \\
\multicolumn{1}{c}{} & \multicolumn{1}{l|}{factor 2} & {\bf 10} & 0 & 0 \\
\multicolumn{1}{c}{} & \multicolumn{1}{l|}{factor 3} & {\bf 10} & 0 & 0 \\ \midrule
\multirow{5}{*}{2D wavy Reaching} & \multicolumn{1}{l|}{factor 1} & 3 & 0 & {\bf 7} \\
 & \multicolumn{1}{l|}{factor 2} & {\bf 8} & 0 & 2 \\
 & \multicolumn{1}{l|}{factor 3} & {\bf 8} & 0 & 2 \\
 & \multicolumn{1}{l|}{factor 4} & {\bf 9} & 1 & 0 \\
 & \multicolumn{1}{l|}{factor 5} & {\bf 9} & 0 & 1 \\ \bottomrule
\end{tabular}
\end{table}

We expect the ladder network can disentangle representations at different levels of abstraction. In this section, we evaluate the factor extracted in each ladder by using 2D Reaching and 2D Wavy Reaching. Table \ref{tab:ladder_level} shows the counting index of the latent variable with the highest mutual information in each ladder network. In Table \ref{tab:ladder_level}, the rows represent factors and columns represent the index of the ladder networks. Factor 1 (goal left/goal right) in 2D Reaching and Factor 1 (goal position) in 2D Wavy Reaching were extracted the most frequently in the latent variable in the 3rd ladder. Since the latent variables have eight dimensions for the 1st ladder, four dimensions for the 2nd ladder, and two dimensions for the 3rd ladder, the 3rd ladder should be the least frequent when factors are randomly entered for each z. Long-term and short-term factors are clear in 2D Wavy Reaching. In 2D Wavy Reaching, there is a distinct difference between factors of long and short time dependency. The "goal position" is the factor that affects the entire trajectory, and the other factors affect half the trajectory (Fig. \ref{fig:2dwave_z2x}). In these experiments, the goal of the trajectory that affects the entire trajectory tended to be expressed in the 3rd ladder. In both datasets, only factor 1 represents goal positions while the others represent the shape of the trajectories. Since factor 1 has a different abstraction level from others, it and the others result in different ladders, e.g., ladder 3 and others.

\subsection{Sprites dataset}
\label{videodataset}

\begin{figure}
\centering
\begin{minipage}{0.49\hsize}
\includegraphics[width=\linewidth]{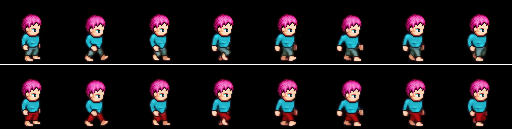}
\subcaption{pant color}
\label{sprite_0_4}
\end{minipage}
\begin{minipage}{0.49\hsize}
\includegraphics[width=\linewidth]{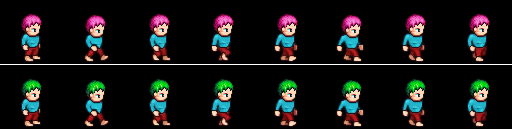}
\subcaption{hair color}
\label{sprite_1_0}
\end{minipage}
\begin{minipage}{0.49\hsize}
\includegraphics[width=\linewidth]{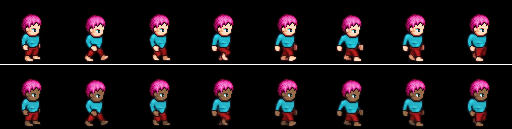}
\subcaption{skin color}
\label{sprite_1_1}
\end{minipage}
\begin{minipage}{0.49\hsize}
\includegraphics[width=\linewidth]{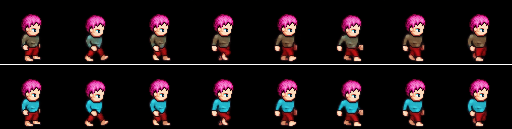}
\subcaption{shirt color}
\label{sprite_0_6}
\end{minipage}
\begin{minipage}{0.49\hsize}
\includegraphics[width=\linewidth]{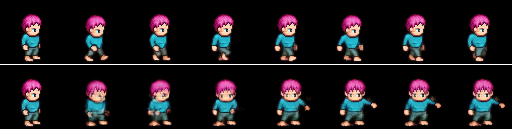}
\subcaption{motion}
\label{sprite_0_3}
\end{minipage}
\begin{minipage}{0.49\hsize}
\includegraphics[width=\linewidth]{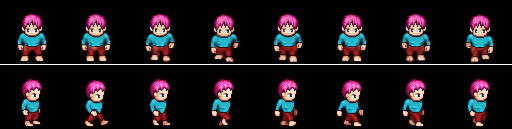}
\subcaption{direction of character}
\label{sprite_0_5}
\end{minipage}
\caption{Visualization of latent traversal of FAVAE. Horizontal axis represents sequence and vertical axis represents differences in $z$.}
\label{fig:dsprite_latenttraversal}
\end{figure}

To evaluate the effectiveness of a video dataset, we trained FAVAE with the Sprites dataset, which was used in~\cite{disney}. This dataset contains $3\times64\times64$ RGB video data with sequential length $= 8$ and consists of static and dynamic factors. Note that motions are not created with the combination of dynamic factors, and each motion exists individually (Dataset detail is explained in Github\footnotemark[2]). We executed disentangled representation learning by using FAVAE with $\beta = 20$, $C = [20, 10, 5]$, and network architecture used for this training is explained in Github\footnotemark[2]. Figure \ref{fig:dsprite_latenttraversal} shows the results of latent traversal, and we use two $z$ values between $-3$ and $3$.  The latent variables in the 1st ladder extract expressions of motion (4th $z$ in 1st ladder), pant color (5th $z$ in 1st ladder), direction of character (6th $z$ in 1st ladder) and shirt color (7th $z$ in 1st ladder). The latent variables in the 2nd ladder extract expressions of hair color (1st $z$ in 2nd ladder) and skin color (2nd $z$ in 2nd ladder). FAVAE can extract the disentangled representations between static and dynamic factors in high dimension datasets. 

\section{Summary and Future Work}
FAVAE learns disentangled and interpretable representations via the information bottleneck from sequential data. The experiments using three sequential datasets demonstrated that it can learn disentangled representations. Future work includes extending the time convolution part to a sequence-to-sequence model~\cite{seq2seq} and applying the model to actions of reinforcement learning to reduce the pattern of actions. 

\bibliographystyle{unsrt} 
\bibliography{favae}

\begin{thebibliography}{10}

\bibitem{Bengio:2013:RLR:2498740.2498889}
Yoshua Bengio, Aaron Courville, and Pascal Vincent.
\newblock Representation learning: A review and new perspectives.
\newblock {\em IEEE transactions on pattern analysis and machine intelligence},
  35(8):1798--1828, 2013.

\bibitem{vae}
Diederik~P Kingma and Max Welling.
\newblock Auto-encoding variational {B}ayes.
\newblock {\em arXiv preprint arXiv:1312.6114}, 2013.

\bibitem{gan}
Ian Goodfellow, Jean Pouget-Abadie, Mehdi Mirza, Bing Xu, David Warde-Farley,
  Sherjil Ozair, Aaron Courville, and Yoshua Bengio.
\newblock Generative adversarial nets.
\newblock In {\em Advances in neural information processing systems}, pages
  2672--2680, 2014.

\bibitem{celeba}
Ziwei Liu, Ping Luo, Xiaogang Wang, and Xiaoou Tang.
\newblock Deep learning face attributes in the wild.
\newblock In {\em Proceedings of the IEEE International Conference on Computer
  Vision}, pages 3730--3738, 2015.

\bibitem{beta-vae1}
Irina Higgins, Loic Matthey, Xavier Glorot, Arka Pal, Benigno Uria, Charles
  Blundell, Shakir Mohamed, and Alexander Lerchner.
\newblock Early visual concept learning with unsupervised deep learning.
\newblock {\em arXiv preprint arXiv:1606.05579}, 2016.

\bibitem{deepvariationalinformationbottleneck}
Alexander~A Alemi, Ian Fischer, Joshua~V Dillon, and Kevin Murphy.
\newblock Deep variational information bottleneck.
\newblock {\em arXiv preprint arXiv:1612.00410}, 2016.

\bibitem{beta-vae2}
Irina Higgins, Loic Matthey, Arka Pal, Christopher Burgess, Xavier Glorot,
  Matthew Botvinick, Shakir Mohamed, and Alexander Lerchner.
\newblock {beta-VAE}: Learning basic visual concepts with a constrained
  variational framework.
\newblock 2016.

\bibitem{infogan}
Xi~Chen, Yan Duan, Rein Houthooft, John Schulman, Ilya Sutskever, and Pieter
  Abbeel.
\newblock {InfoGan}: Interpretable representation learning by information
  maximizing generative adversarial nets.
\newblock In {\em Advances in neural information processing systems}, pages
  2172--2180, 2016.

\bibitem{dqn-macro-action}
Ishan~P Durugkar, Clemens Rosenbaum, Stefan Dernbach, and Sridhar Mahadevan.
\newblock Deep reinforcement learning with macro-actions.
\newblock {\em arXiv preprint arXiv:1606.04615}, 2016.

\bibitem{fhvae}
Wei-Ning Hsu, Yu~Zhang, and James Glass.
\newblock Unsupervised learning of disentangled and interpretable
  representations from sequential data.
\newblock In {\em Advances in neural information processing systems}, pages
  1878--1889, 2017.

\bibitem{disney}
Yingzhen Li and Stephan Mandt.
\newblock A deep generative model for disentangled representations of
  sequential data.
\newblock {\em arXiv preprint arXiv:1803.02991}, 2018.

\bibitem{isolating}
Tian~Qi Chen, Xuechen Li, Roger Grosse, and David Duvenaud.
\newblock Isolating sources of disentanglement in variational autoencoders.
\newblock {\em arXiv preprint arXiv:1802.04942}, 2018.

\bibitem{factorvae}
Hyunjik Kim and Andriy Mnih.
\newblock Disentangling by factorising.
\newblock {\em arXiv preprint arXiv:1802.05983}, 2018.

\bibitem{matthew}
Matthew~D Hoffman and Matthew~J Johnson.
\newblock {ELBO} surgery: yet another way to carve up the variational evidence
  lower bound.
\newblock In {\em Workshop in Advances in Approximate Bayesian Inference,
  NIPS}, 2016.

\bibitem{totalorrelation}
Satosi Watanabe.
\newblock Information theoretical analysis of multivariate correlation.
\newblock {\em IBM Journal of research and development}, 4(1):66--82, 1960.

\bibitem{understand-beta-vae}
Christopher~P Burgess, Irina Higgins, Arka Pal, Loic Matthey, Nick Watters,
  Guillaume Desjardins, and Alexander Lerchner.
\newblock Understanding disentangling in $\beta$-{VAE}.
\newblock {\em arXiv preprint arXiv:1804.03599}, 2018.

\bibitem{informationbottleneck}
Naftali Tishby, Fernando~C Pereira, and William Bialek.
\newblock The information bottleneck method.
\newblock {\em arXiv preprint physics/0004057}, 2000.

\bibitem{informationdropout}
Alessandro Achille and Stefano Soatto.
\newblock Information dropout: Learning optimal representations through noisy
  computation.
\newblock {\em IEEE Transactions on Pattern Analysis and Machine Intelligence},
  2018.

\bibitem{informationbottleneckgaussinan}
Gal Chechik, Amir Globerson, Naftali Tishby, and Yair Weiss.
\newblock Information bottleneck for {G}aussian variables.
\newblock {\em Journal of machine learning research}, 6(Jan):165--188, 2005.

\bibitem{VRNN}
Junyoung Chung, Kyle Kastner, Laurent Dinh, Kratarth Goel, Aaron~C Courville,
  and Yoshua Bengio.
\newblock A recurrent latent variable model for sequential data.
\newblock In {\em Advances in neural information processing systems}, pages
  2980--2988, 2015.

\bibitem{SRNN}
Marco Fraccaro, S{\o}ren~Kaae S{\o}nderby, Ulrich Paquet, and Ole Winther.
\newblock Sequential neural models with stochastic layers.
\newblock In {\em Advances in neural information processing systems}, pages
  2199--2207, 2016.

\bibitem{STORN}
Justin Bayer and Christian Osendorfer.
\newblock Learning stochastic recurrent networks.
\newblock {\em arXiv preprint arXiv:1411.7610}, 2014.

\bibitem{DRAW}
Karol Gregor, Ivo Danihelka, Alex Graves, Danilo~Jimenez Rezende, and Daan
  Wierstra.
\newblock {DRAW}: A recurrent neural network for image generation.
\newblock {\em arXiv preprint arXiv:1502.04623}, 2015.

\bibitem{vlae}
Shengjia Zhao, Jiaming Song, and Stefano Ermon.
\newblock Learning hierarchical features from generative models.
\newblock {\em arXiv preprint arXiv:1702.08396}, 2017.

\bibitem{lstm}
Sepp Hochreiter and J{\"u}rgen Schmidhuber.
\newblock Long short-term memory.
\newblock {\em Neural computation}, 9(8):1735--1780, 1997.

\bibitem{gru}
Junyoung Chung, Caglar Gulcehre, KyungHyun Cho, and Yoshua Bengio.
\newblock Empirical evaluation of gated recurrent neural networks on sequence
  modeling.
\newblock {\em arXiv preprint arXiv:1412.3555}, 2014.

\bibitem{qrnn}
James Bradbury, Stephen Merity, Caiming Xiong, and Richard Socher.
\newblock Quasi-recurrent neural networks.
\newblock {\em arXiv preprint arXiv:1611.01576}, 2016.

\bibitem{batchnorm}
Geoffrey~E Hinton, Nitish Srivastava, Alex Krizhevsky, Ilya Sutskever, and
  Ruslan~R Salakhutdinov.
\newblock Improving neural networks by preventing co-adaptation of feature
  detectors.
\newblock {\em arXiv preprint arXiv:1207.0580}, 2012.

\bibitem{eastwood2018a}
Cian Eastwood and Christopher~KI Williams.
\newblock A framework for the quantitative evaluation of disentangled
  representations.
\newblock 2018.

\bibitem{timit}
John~S Garofolo, Lori~F Lamel, William~M Fisher, Jonathan~G Fiscus, and David~S
  Pallett.
\newblock {DARPA TIMIT acoustic-phonetic continous speech corpus CD-ROM. NIST
  speech disc 1-1.1}.
\newblock {\em NASA STI/Recon technical report n}, 93, 1993.

\bibitem{aurora}
David Pearce and J~Picone.
\newblock {Aurora working group: DSR front end LVCSR evaluation AU/384/02}.
\newblock {\em Inst. for Signal \& Inform. Process., Mississippi State Univ.,
  Tech. Rep}, 2002.

\bibitem{adam}
Diederik~P Kingma and Jimmy Ba.
\newblock Adam: A method for stochastic optimization.
\newblock {\em arXiv preprint arXiv:1412.6980}, 2014.

\bibitem{seq2seq}
Ilya Sutskever, Oriol Vinyals, and Quoc~V Le.
\newblock Sequence to sequence learning with neural networks.
\newblock In {\em Advances in neural information processing systems}, pages
  3104--3112, 2014.

\end{thebibliography}

\end{document}